# Enabling Edge Cloud Intelligence for Activity Learning in Smart Home

Bing Huang
School of Computer Science
The University of Sydney
Sydney, Australia
bing.huang@sydney.edu.au

Athman Bouguettaya
School of Computer Science
The University of Sydney
Sydney, Australia
athman.bouguettaya@sydney.edu.au

Hai Dong
School of Science
RMIT University
Sydney, Australia
hai.dong@rmit.edu.au

*Abstract*—We propose a novel activity learning framework based on Edge Cloud architecture for the purpose of recognizing and predicting human activities. Although activity recognition has been vastly studied by many researchers, the temporal features that constitute an activity, which can provide useful insights for activity models, have not been exploited to their full potentials by mining algorithms. In this paper, we utilize temporal features for activity recognition and prediction in a single smart home setting. We discover activity patterns and temporal relations such as the order of activities from real data to develop a prompting system. Analysis of real data collected from smart homes was used to validate the proposed method.

## I. INTRODUCTION

The Internet is evolving from connecting computers to connect surrounding things of people's living space (i.e., Internet of Things)[25]. With the emergence of various IoT devices such as switch sensors, infrared motion sensors, pressure sensor, wearable sensors, accelerators, physical things are enabled to be connected and augmented with capabilities such as sensing, actuating, and communication [30]. In this regard, connected things with these augmented capabilities are referred as *IoT services*[23]. Common things such as a connected air-conditioner can be represented as an airconditioner service.

An important application domain for IoT is the smart home. A smart home can be considered as any regular home which has been augmented with various types of IoT services [23]. An activity is defined as a sequence of actions exerted on IoT services [57]. Taking "making coffee" as an example, the resident needs to walk to the kitchen, open the fridge, take out the milk, take a coffee cup, turn on a coffee maker, and turn off the coffee maker. The resident performs a variety of daily activities via interacting with various IoT services. As a result, there will be large volumes of data generated by interacting with IoT services. The availability of large volumes of data will spur opportunities to extract insights and knowledge for supporting daily activities. One such opportunity is to improve daily life *convenience*. Convenience is underpinned by the principle of the *least effort*, i.e. the premise that humans would usually want to achieve goals with the least cognitive and physical efforts [59]. In the smart home setting, convenience could be achieved through reducing residents' interactions with IoT services. We use a motivating scenario to depict the prompting service system which monitors and tracks human activity in order to deliver life convenience. Motivating scenario. Suppose Sarah lives alone in the smart home. Daily life things such a TV, an airconditioner are connected one the Internet and represented as IoT services. She performs daily activities via interacting with these IoT services, which is monitored and analysed. As a result, her ongoing activity at a particular location is known by the IoT-based prompting system. For example, Sarah is waking up in the morning and the system know this activity. A voice message is prompted to ask whether she wants to turn on the heater preheat the bathroom. Sarah answers "Yes" to this prompts. The heater is started to heating the bathroom. Therefore, when she goes to the bathroom to take a shower, it is already warm and she does not need to wait. After finishing taking a shower, she goes to the kitchen to cook breakfast. When she turns on a stove, a voice message is prompted to remind her to turn on the ventilator. She answers "Yes" to this prompt and the ventilator is turned on automatically.

In the motivating scenario, *activity learning* is a key task for realizing convenience. Activity learning generally refers to the learning and understanding of the observed human activities [10]. It involves three main subtasks: (1) *activity discovery*, which focuses on finding unknown activity patterns from unlabeled IoT service usage data [26]; (2) *activity recognition*, which aims to use machine learning techniques to map a sequence of IoT service usage data to a corresponding activity label [11]; and (3) *activity prediction*, which is aimed at inferring which activity will be performed in the future [10]. *In this paper, we focus on designing an activity learning framework based on Edge Cloud architecture for the purpose of fully realizing the potential of smart homes in terms of delivering daily life convenience.*

There is much research on building ambient intelligent smart home environment ranging from video-based approaches to sensor-based approaches. Video-based approaches leverage on video cameras as passive sensors to recognize people's actions from video sequences [7]. They mainly focus on recognizing simple actions such as walking, standing, and waving hands.

However, in a smart home setting, video-based approaches have privacy and security issues. Sensor-based approaches in activity recognition are gaining more popularity.

Many activity recognition algorithms are supervised, i.e., they rely on labeled data for training [47]. A decision tree is trained for recognizing simple activities such as walking and standing [3]. Naive Bayesian classifiers are used in [53] to recognize complex activities such as toileting and bathing from sensor data collected from daily objects [52]. Other sophisticated supervised approaches include support vector machine, recurrent neural networks, and hidden Markov model etc [42]. Supervised methods do not scale well because sensor data need to be annotated manually, which is a very time consuming and laborious task. Recently, a few unsupervised approaches have been proposed to address the problem of activity recognition. In [48], a frequent sequence pattern mining approach is proposed to extract activity patterns from sensor data. The hidden Markov model is constructed these activity patterns. In [15], a pattern mining approach is proposed to discover emerging patterns from sensor data. The discovered patterns are employed to build a classifier to recognize both simple and complex activities. However, these unsupervised approaches focus on recognizing ongoing activities and do not consider predicting the next action within a specific activity.

As previous discussion, the state-of-the-art studies conducted in smart homes are from a *technique* perspective. As a result, many sophisticated supervised and unsupervised techniques are devised to recognize daily activities ranging from simple activities (e.g., "walking" ) to complex activities (e.g.,"taking a shower"). Indeed, these techniques are important research topics in smart homes. However, to the best of knowledge, there is not much research on designing an intelligent system from a *value* perspective. Values refers to relative economic worth or utility, and is the basis of human endeavors aimed at efficient use of resources based on concerns such as least cost [33]. We believe that *convenience* is an important concern or value in the smart home domain. In this paper, we take a value based perspective in terms of *convenience* to design an activity learning framework based on Edge Cloud architecture. In a nutshell, contributions are as follows:

- We formulate convenience delivering as an activity learning problem. In particular, we propose a unified activity learning framework to recognize ongoing activities and predict the next activity.
- The activity discovery component aims to extract two types of patterns: activity patterns and temporal patterns. We design a new agglomerative algorithm considering time, location, and feature information to cluster activities. Each cluster is transformed into a corresponding activity pattern. We also propose a novel algorithm for discovering temporal patterns among activities
- We design a rule-based approach to segment the boundary of any two adjacent activities. We also develop an algorithm to recognize ongoing activities and predict the next activity.
- We conduct a series of experiments to validate the effectiveness and efficiency of our proposed activity learning framework.

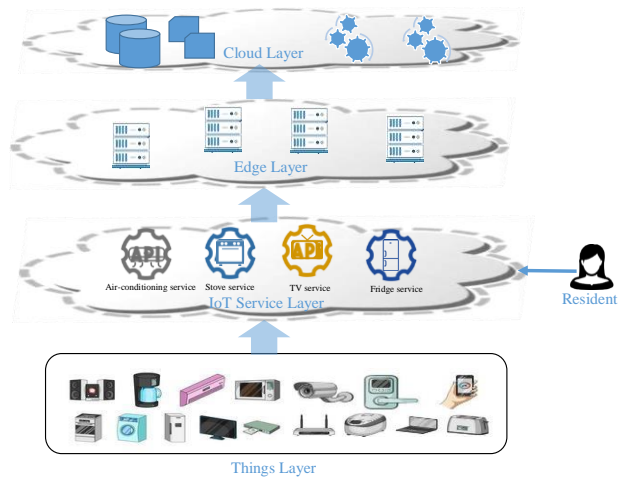

Fig. 1. The architecture of the smart home system

## II. ACTIVITY LEARNING FRAMEWORK

In this section, we first overview of the activity learning framework based on Edge Cloud architecture, followed by a brief description of the key components. We consider a four-layer system architecture for activity learning (see Fig.1): *Thing*, *IoT service*, *Edge*, and *Cloud*. The things layer consists of all artifacts (aka things) that are internet enabled. These things are equipped with IoT devices such as sensors and actuators. These things augmented with IoT devices are wrapped as IoT Services [23]. The usage data of IoT services is pushed into the Cloud for further processing. The Cloud layer takes care of: (i) collecting and managing raw IoT service usage data, (ii) training an activity model, and (iii) managing Edge nodes. We assume that each smart home comes with an Edge server to allow low levels of latency for decision making [12]. The trained activity model is deployed on the Edge to recognize activities and predict next action within an activity. The Edge layer is responsible for: (i) segmenting streaming service usage data into activity sequences, (ii) recognizing activities and predicting next action based on the trained activity model. We will provide more technical details on data segmentation and activity model training in section III. The trained activity model serves the basis for recognizing ongoing activities and predicting the next action. The recognized activity (i.e., an activity label) and predicted action can be harnessed to enable the IoT-based prompting system (section IV).

## III. ACTIVITY DISCOVERY

The activity discovery component focuses on extracting two types of patterns: activity patterns and temporal patterns. We first provide some basic preliminaries and formally define activity patterns and temporal patterns as follows.

Definition 1: IoT Service. An IoT service $S_i$ is represented by a tuple $S_i = <F_i, Q_i>$ where $F_i$ is a set of functionalities and $Q_i$ is a set of qualities offered by the IoT service. For example, the functionality of a light service is to provide illumination and its qualities may include the brightness.

Definition 2: IoT Service event. The usage of an IoT service is recorded as an event. It is denoted as $e_i = <s_i, \alpha_i, t_i, l_i)>$ where $s_i$ is the ID of the IoT service $S_i$, $\alpha_i$ is an event type, $t_i$ is a time stamp, and $l_i$ is a location. For example, the light is turned on at 3pm in the bedroom, which is represented as <L100, ON, 3pm, bedroom> (i.e., L100 is the ID of the light and "ON" is an event type.).

Definition 3: IoT Service event sequence. An IoT service event sequence consists of multiple events that are ordered by their time stamps. It is defined as $E = \{e_1, e_2,...,e_n\}$ where any two adjacent events satisfy $t_i \le t_{i+1}$.

*A. Activity pattern discovery*

As previous discussion, we define an activity as a set of actions. For example, the "making coffee" activity involves a set of actions {*Open the fridge, take the milk, close the fridge, take the coffee cup, turn on the coffee maker, turn off the coffee maker*}. Each IoT service is attached with a sensor to monitor its usage. As a result, each action could be inferred from IoT service event data. For example, *(I001, ON)* and *(I001, OFF)* represent the "open the fridge" and "close the fridge", respectively. In this regard, an activity could be represented as a set of event types. When each event type in an action set is associated with a time stamp and a location, we call such set as an *occurrence* of the activity. We formalize the activity and the activity occurrence as follows.

Definition 4: Activity. An activity is a set of actions, which is represented as $A_i = \{<s_1, \alpha_1>, <s_2, \alpha_2>,...,<s_n, \alpha_n>\}$ where $<s_i, \alpha_i>$ is a tuple of the IoT service ID $s_i$ and the event type $\alpha_i$. For the purpose of simplicity, we use the symbol $a_i$ (called event type) to replace $<s_i, \alpha_i>$. Therefore, $A_i$ is simplified as $A_i = \{a_1, a_2,...,a_n\}$.

Definition 5: Activity occurrence. An activity occurrence $AO_i$ describes the time and location of the activity $A_i$. It is represented as $AO_i = <SID, E_i>$ where $SID$ is a unique identifier and $E_i = \{<a_1, t_1, l_1>, <a_2, t_2, l_2>,...,<a_m, t_m, l_m>\}$ is an event sequence.

It is worthy noting that an activity could be performed in various manners. For example, consider the "making coffee" activity. The resident may not perform this activity in exactly the same way each time. Some of the steps might be changed or varied in order and some IoT services may not be used some time. As a result, there are different activity occurrences for the same activity. The activity pattern should capture such variation and sequential features of an activity. We use Hidden Markov Model(HMM) to represent an activity pattern [50]. HMM is a probabilistic graphic model in which the observations are generated by a stochastic process that is not observable [13].

Definition 6: Activity pattern. An activity pattern $\lambda$ is formalized as an HMM as follows:

$$\lambda = (A, B, o) \quad (1)$$

$A = [a_{ij}]$ is a transition probability matrix, storing the probability of state $j$ following state $i$ where:

$$a_{ij} = P(q_t = s_j | q_{t-1} = s_i) \quad (2)$$

$B = [b_{ik}]$ is an emission probability matrix, storing the probability of observation $k$ being produced from the state $j$ where:

$$b_{ik} = P(x_t = v_k | q_t = s_i) \quad (3)$$

$o = [o_i]$ is an initial probability matrix, storing the probability distribution over initial states where:

$$o_i = P(q_1 = s_i) \quad (4)$$

As previous discussion, for each activity, it could be performed in various ways, resulting multiple similar activity occurrences for the same activity. In order to find activity patterns, we design an aggolerative clustering approach to group similar activity occurrences into a cluster. Therefore, each cluster corresponds to an activity activity. The essential problem of clustering is to design a similarity measure. In this paper, we compute the similarity *sim* between activity occurrences using the following equation:

$$sim(AO_i, AO_j) = \Upsilon_l + \Upsilon_t + \Upsilon_a \quad (5)$$

where $\Upsilon_l$ refers to location similarity (i.e., if two activities occur in the same location); $\Upsilon_t$ is the time similarity; and $\Upsilon_a$ refers to structure similarity in terms of involved event types.

$$\Upsilon_l = \begin{cases} 1, & AO_i.loc = AO_j.loc \\ 0 & otherwise \end{cases} \quad (6)$$

We employ the similarity measure proposed in [23] to compute the time similarity between $AO_i$ and $AO_j$. Specifically, we utilize a function $f_i$ with respect to $t$ to map the temporal aspect of $AO_i$. $f_i$ is formalized as follows.

$$f_i(t) = \begin{cases} 1, & t \in [st_i, et_i] \\ 0 & otherwise \end{cases} \quad (7)$$

Then we have a set of functions $f_1, f_2,...,f_n$ corresponding to each activity occurrence. The time similarity $\Upsilon_t$ is formalized as: beginequation

$$\Upsilon_t = \frac{\int_{t_1}^{t_4} \sum_{i=1}^{2} f_i(t)\, dt}{(t_4 - t_1) \cdot 2} \quad (8)$$

where $t_1$ and $t_2$ are the the start time and end time for the activity occurrence $AO_i$, respectively. $t_3$ and $t_4$ are the start time and end time for $AO_j$, respectively. $\Upsilon_t$ ranges from 0 to 1 and the larger the value, the more similar between two activity occurrences in terms of time. For example, the "preparing dinner" starts at 18:00 and ends at 19:00 and "washing clothes" is performed during 18:40 and 19:20. The time similarity between the two activities are calculated as $\frac{(18:40-18:00)+(19:00-18:40)\cdot 2+(19:20-19:00)}{(19:20-18:00)\cdot 2}$ = 0.625. We use Jaccard similarity to compute the structure similarity in terms of event types. $\Upsilon_a$ is defined as follows:

$$\Upsilon_a = \frac{|A_i \cap A_j|}{|A_i \cup A_j|} \quad (9)$$

where $A_i$ and $A_j$ are the sets of event types involved in the activity occurrences $AO_i$ and $AO_j$, respectively.

Next, we apply the aggolermerative hierarchical clustering algorithm to cluster the set of activity occurrences into multiple groups based on the similarity measure in Equation 5. Each cluster is associated with a corresponding activity pattern. The classical aggolermerative hierarchical clustering algorithm uses a bottom-up merging strategy [20]. It typically starts with treating each data object as initial clusters and iteratively merges close clusters into larger clusters, until all data objects are in a single cluster. The single cluster is the hierarchy's root. Similar with the classical agglomerative hierarchical clustering approach, we initialize each data object(i.e., activity occurrence) as a cluster and iteratively merge close clusters into larger ones. Instead of clustering up to the point of reaching the root cluster, we continue the clustering until the distance between the two clusters rises above a distance threshold $\rho$. By doing do, we could find a set of clusters at the highest level of the hierarchy. Our clustering algorithm is shown in Algorithm 1. First of all, we compute the similarity matrix $m$ based on Equation 5. Then we can have a distance matrix $M$ (Algorithm 1 line 3). We use average linkage method to compute the distance between clusters in Equation 10.

$$D(c_i, c_j) = T_{ij}/(N_i * N_j) \quad (10)$$

where $T_{ij}$ is the sume of all pairwise distances between cluster $c_i$ and cluster $c_j$. $N_i$ and $N_j$ are the sizes of the cluster $c_i$ and cluster $c_j$, respectively. Then, we merge two clusters into a larger one if their distance is less than the threshold $\rho$ (algorithm line 6). We continue clustering until the distance between two clusters is no less than $\rho$. Therefore, we have a set of clusters and each cluster corresponds to an activity. For each cluster, we collect all involved event types and compute the occurrence probability distribution and conditional probability distribution for all event types. The occurrence probability distribution is mapped to the emission probability matrix and the conditional probability distribution is mapped to the transition probability matrix. As a result, the HMM or the activity pattern is constructed automatically for each cluster.

*B. Temporal pattern discovery*

It is an essential task to extract temporal patterns for activities. The temporal patterns refer to the temporal relationship patterns among activities. These temporal patterns describes how residents allocate time across activities and how they transit from performing one activity to another. The discovered temporal patterns serve the basis for predicting the next activity. However, it is non-trivial to extract temporal relationships among activities.

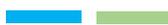

| Temporal relations | Formalization | Pictorial example |
|---|---|---|
| $S_i$ before $S_j$ | $et_i < st_j$ | 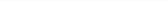 |
| $S_i$ overlap $S_j$ | $st_i < st_j < et_i < et_j$ | |
| $S_i$ equal $S_j$ | $st_i = st_j$ and $et_i = et_j$ | |
| $S_i$ start-by $S_j$ | $st_i = st_j$ and $et_i < et_j$ | |
| $S_i$ finish $S_j$ | $st_i > st_j$ and $et_i = et_j$ | |
| $S_i$ meet $S_j$ | $et_i = st_j$ | 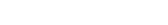 |
| $S_i$ during $S_j$ | $st_j < st_i < et_i < et_j$ | |

Fig. 2. Temporal relationships among activities

are much more complex than point-based data. Thus a new model that can capture the complex relationships among time intervals is required. Allen's temporal logic is a commonly used model to describe the complex temporal relationships among time intervals. In this paper, we adapt Allen's temporal logic to model the temporal relationships among activities as shown in Fig.2. In this figure, we assume the start time and end time for the activity $S_i$ are $st_i$ and $et_i$ (resp. [$st_j$,$et_j$] for the activity $S_j$), respectively. We formally define temporal patterns as follows.

*A temporal pattern is a collection of activities that occur frequently together in a particular temporal relationship*. An example of temporal pattern is shown in Fig? The key problem for modeling temporal patterns is to represent temporal relations in an unambiguous way. In this paper, we adopt the endpoint representation method proposed in [9] to describe complex temporal relations among time intervals. Specifically, for the activity $S_i$ starting at $st_i$ and finishing at $et_i$, it is denoted as ($<s^+_i,st_i>,<s^-_i,et_i>$), where $<s^+_i,st_i>$ (resp. $<s^-_i,st_i>$) is the start state (resp. end state) and $s^*$ (*=+ or -) is a state symbol. Therefore, a collection of $n$ activity occurrences are represented as

$$S = \{<(s^+_1,st_1),(s^-_1,et_1)>,...,<(s^+_n,st_n),(s^-_n,et_n)>\}.$$

By ordering all elements $s_i^*$ (* can be + or -) in $S$ in a non-decreasing order based on its associated time information $st_i$(or $et_i$), we can transform $S$ into the following representation

$$S =< Seq,T >= \begin{Bmatrix} \alpha_1 & ... & \alpha_i & ... & \alpha_{2n} \\ t_1 & ... & t_i & ... & t_{2n} \end{Bmatrix},$$

where $Seq = \{\alpha_1...\alpha_i...\alpha_{2n}\}$ is a state symbol sequence and $\alpha_i = s_j^*$ (* can be + or -), $T = \{t_1...t_i...t_{2n}\}$ is the time information. For example, the composite IoT service in Fig?

can be represented as $\begin{Bmatrix} s^+_2 & s^+_1 & s^+_3 & s^-_1 & s^-_3 & s^-_2 \\ 48 & 50 & 58 & 65 & 70 & 75 \end{Bmatrix}$. After transforming the temporal relations between time intervals into the endpoint representation, we have a state symbol sequence. As a result, the problem of temporal pattern mining is transformed as the well-known frequent sequence pattern mining problem. We formalize temporal patterns as follows.

Definition 7: Temporal pattern. Given a collection of activity occurrences in endpoint representation, a temporal pattern is a sequence of state symbols that occurs frequently. It is represented as $TP =< Seq,sup >$ where:

- $Seq = \{\alpha_1...\alpha_i...\alpha_{2m}\}$ is the state symbol sequence.
- $sup(Seq)$ is the support $Seq$. It is the total number of occurrences for a particular activity. Let us assume $Seq = \{\alpha_1...\alpha_i...\alpha_{2m}$ and $Seq' = \{\alpha_1...\alpha_i...\alpha_{2n}$ be two state symbol sequences. $Seq$ is referred to as a sub-sequence of $Seq'$, denoted as $Seq \sqsubseteq Seq'$ if there exist integers
  $1 \le k_1 \le k_2...k_n \le k_{2m}$ such that $\alpha_1 \subseteq \alpha'_{k_1}$, $\alpha_2 \subseteq \alpha_{k'_2},..., \alpha_n \subseteq \alpha_{k'_n}$.

Given a set of state symbol sequences in DB, the tuple ($sid,Seq'$) (i.e., $sid$ is a sequence ID and $Seq'$ is the state symbol sequence) is said to contain a sub-sequence $Seq$ if $Seq \sqsubseteq Seq'$. The support $sup$ for $Seq$ in DB, denoted as $sup(Seq)$ is the number of tuples containing $Seq$. $sup(Seq)$ can be formalized as follows.

$$sup(Seq) = |\{(sid,Seq) \in DB | Seq \sqsubseteq Seq'\}| \qquad (11)$$

We design the algorithm TPMiner (Temporal Pattern Miner) to extract temporal patterns from the state symbol sequences database. TPMiner adopts a divide-and-conquer, patterngrowth principle from the algorithm Prefixspan [44] as follows: the database are recursively projected into a set of smaller *projected databases* based on the current temporal patterns. Temporal patterns are then grown by searching the smaller projected databases instead of searching the whole database. We define the key concept *projected database* and explain the pattern mining process in details as follows.

Definition 8: Projected database. Let $p$ be a temporal pattern in the database $DB$. The $p$-projected database, denoted as $DB|_p$, is the set of suffixes of state symbol sequences in $DB$ with regard to the prefix $p$.

The mining process consists of three steps. In the first step, TPMiner searches a collection of patterns with one length, resulting in a 1-length temporal patterns set. Next, for each 1length pattern, associated projected databases are constructed. Then, the length of each pattern is grown through searching its associated projected databases. The three steps are detailed as follows.

*Step 1. Search the set of 1-length temporal patterns $L_1$.*

---

**Algorithm 1** Agglomerative hierarchical clustering

**Input:** $D = \{AO_1, AO_2, ..., AO_n\}$(a set of activity occurrences data), $\rho$(distance threshold)
**Output:** $C = \{c_1, c_2, ..., c_k\}$(a set of clusters)
1: Set each data object as a initial cluster.
2: Compute similarity matrix $m$ by Equation 5.
3: $M = 1-m$ // Compute distance matrix.
4: **repeat**
5:     $distance = D(c_i, c_j) \forall c_i, c_j \in C$
6:     **if** $distance < \rho$ **then**
7:         merge($c_i, c_j$)
8:     **end if**
9:     Update M;
10: **until** $distance \ge \rho$
11: **return** $C$

---

**Algorithm 2** TPMiner algorithm

**Input:** $DB$(A set of state symbol sequences), $minsup$(support threshold)
**Output:** $patternset$( a set of temporal patterns)
1: $L_1 = $ find_1-length_pattern($DB_{r_i}$, $minsup$ );
2: $Prefixspan(\alpha, l, DB_{r_i}|_\alpha)$ //$\alpha$ is a temporal pattern, $l$ is the length of $\alpha$ ($l \ge 1$), $DB_{r_i}|_\alpha$ is the $\alpha$-projected database.
3: Scan $DB_{r_i}|_\alpha$ once, find the set of frequent state symbols $\{e_1, e_2...e_n\}$;
4: **for each** frequent state symbol $e_i$ **do**
5:     $\alpha' = \alpha + e_i$; //append $e_i$ to $\alpha$ to form the new temporal pattern $\alpha'$.
6: **for each** $\alpha'$ **do**
7:     add $\alpha'$ to the temporal pattern set $patternset$;
8:     construct $\alpha'$-projected database $DB_{r_i}|_{\alpha'}$, and call $Prefixspan$ ($\alpha', l+1, DB_{r_i}|_{\alpha'}$);
9: **return** $patternset$;

counting the frequency of each state symbol pairs and removes those with low frequency. For instance, if we set the *minsup* threshold to be 2, the 1-length patterns whose frequency is more than 2 remain and consist of the 1-length pattern set $L_1$. *Step 2. Construct projected databases for each 1-length temporal pattern*. Suppose $L_1 = \{\alpha_1^1, \alpha_2^1...\alpha_n^1\}$ is the set 1length temporal patterns set. For an element $\alpha_i^1$, an associated projected database $DB|_{\alpha_i^1}$ is constructed. According to the definition 7, $DB|_{\alpha_i^1}$ is the set of suffix event sequences based on its prefix $\alpha_i^1$.

*Step 3. A k-length temporal pattern α is extended to the (k+1)length pattern α' by searching its projected database $DB|_\alpha$ ( $k \geq 1$ )*. Given a prefix $\alpha$ temporal pattern, TPMiner finds the local frequent state symbol pairs through searching its projected database $DB|_\alpha$. The local frequent state symbol pairs constitutes the set $\{e_1, e_2...e_n\}$ and infrequent ones are removed. Then, the frequent state symbol pair $e_i$ is appended to the prefix $\alpha$, resulting a new frequent temporal pattern $\alpha'$ with the length increased by 1. Thus, the (k+1)-length temporal patterns prefixed with $\alpha$ are generated. Note that we consider state symbol pairs and single state symbol in the projected database will not be considered again.

## IV. ACTIVITY RECOGNITION AND PREDICTION

Once the activity patterns and temporal patterns are discovered for the resident, we want to build a model that is able to *recognize* the ongoing activity and *predict* future activities. This will allow the IoT-based smart home environment to track each activity and predict future activities for the purpose of making appropriate prompts. Suppose the resident is performing a particular activity by activating a number of IoT services during a time period and these observations have been recorded as a new IoT service usage sequence. The activity recognition component(i.e., activity recognizer) aims to map the new IoT service usage sequence to an activity label. The activity prediction component(i.e., activity predictor) focuses on inferring future activity labels based on the outcomes produced by the activity recognizer. We discuss the details of the two components as follows.

```
Algorithm 3 Forward algorithm
Input: O = {o₁, o₂, ..., o_T}(A new IoT service usage sequence),
       λ(A HMM model)
Output: P(O|λ)( A probability)
 1: Create a matrix α[N][T]; // N is the number of states in λ and
    T is the length of O.
 2: for i = 1 to i = N do
 3:     α[1][i] = o_i b_i(O_1);
 4: end for
 5: for t = 1 to t = T − 1 do
 6:     for j = 1 to j = N do
 7:         α[t + 1][j] = [∑_{i=1}^{i=N} α[t][i] ∗ α[i][j]] ∗ b_j(O_{t+1})
 8:     end for
 9: end for
10: P(O|λ) = ∑_{i=1}^{i=N} α[T][i];
```

Let us define the new sequence of IoT service usage data as $S = \{s_1, s_2,..., s_n\}$. Given a set of activity class $C = \{c_1, c_2,..., c_{nc}\}$, we want to decide which class $c_i$ the sequence $S$ belongs to. As discussed earlier, we have trained one corresponding HMM $m_i$ for each class $c_i$. As a result, we have a set of models $M = \{m_1, m_2,..., m_{nc}\}$ corresponding to the activity class $C$. The objective of the activity recognizer is to map the sequence $S$ to the mostly likely model $m_*$, that is,

$$m_* = \underset{m_i \in M}{\arg\max} P(S|m_i) \quad (12)$$

It is worth noting that the new IoT service usage sequence data will be first partitioned into segments so that the sequence of IoT service usage observations within a segment come from an activity. Segmenting streaming data is an important research topic and there are many studies related to this topic [41][34][32]. In this paper, the problem of segmenting streaming IoT service usage data is out the scope of our discussion and we assume the usage data has been segmented.

## V. EXPERIMENT

We evaluate the effectiveness and efficiency of our proposed activity learning framework. We conduct a set of experiments on a 3.4 GHZ Intel processor and 8 GB RAM under Windows 10 environment.

### A. Experiment design and Dataset description

We conduct our experiments on the real world dataset that are collected from an old person who lived alone in a smart apartment [53]. There were 84 sensors attached on daily life things such as doors, windows, cabinets, drawers, microwave ovens, refrigerators, stoves, sinks, toilets, showers, light switches, lamps, some containers (e.g water, sugar, and cereal), and electric/electronic appliances (e.g DVDs, stereos, washing machines, dish washers, coffee machines). The locations of these sensors are also provided. The start time and end time of related sensors are recorded when the resident performs activities. Fig.3 shows some samples of the dataset. For example, the "taking medication" is performed from time 4:23:06am to 4:41:32am in the kitchen on the date of 5/3/2003. There are 6 things are involved in this activity including Light switch with sensor ID 75, Door with sensor ID 51, Sink faucet with sensor ID 91 etc. Each thing has a start time and end time. The Light switch identified by sensor 75 starts at 4:30:23am and ends at 4:33:39am. A total of 21 activity types are annotated and recorded. All these activities are performed in a natural setting without any interruption in the daily life of the resident. The data was collected during an approximately two weeks period.

| Taking medication | 5/3/2003 | 4:23:06 | 4:41:32 | Kitchen | | |
|---|---|---|---|---|---|---|
| 75 | 51 | 91 | 96 | 119 | 74 | |
| Light switch | Door | Sink faucet | Sink faucet | Light switch | Refrigerator | |
| 4:30:23 | 4:30:34 | 4:31:29 | 4:31:33 | 4:32:10 | 4:32:40 | |

| 4:33:39 | 19:41:00 | 4:55:53 | 6:11:38 | 20:30:34 | 4:32:47 | |
|---|---|---|---|---|---|---|
| Preparing breakfast | 5/3/2003 | 5:29:51 | 6:24:47 | Kitchen | | |
| 54 | 74 | 75 | 74 | 108 | 84 | 108 |
| Cabinet | Refrigerator | Light switch | Refrigerator | Toaster | Garbage disposal | Toaster |
| 5:36:36 | 5:37:23 | 5:37:55 | 5:38:08 | 5:38:27 | 5:39:54 | 5:40:20 |
| 5:36:38 | 5:37:50 | 5:40:31 | 5:38:24 | 5:38:33 | 6:11:57 | 6:15:37 |

Fig. 3. Samples of datasets

### B. Analysis and evaluation

*1) Results and Analysis on discovering activity patterns.* We implement the algorithm 1 to cluster similar activity occurrences together. We employ two metrics including Bcubed and execution time to evaluate the clustering quality and performance of our clustering algorithm, respectively.

Bcubed is a widely used metric for evaluating the quality of clustering algorithm based on given ground truth [20]. It evaluates the precision and recall for each object in a cluster based on externally given ground truth. The precision of an object indicates how many other objects in the same cluster belong to the same category as the object. The recall of an object reflects how many objects of the same category are assigned to the same cluster. Therefore, BCubed precision is the averaged precision of all objects in the data. The BCubed recall is analogous, replacing cluster with category. We formally define BCubed precision and BCubed recall as well as F1 score as follows.

Let $L(o)$ and $C(o)$ denote the category and the cluster of an object $o$. We can define the *correctness* of the relation between $o$ and $o'$ as:

$$correctness(o, o') = \begin{cases} 1, if L(o) = L(o') \Leftrightarrow C(o) = C(o') \\ 0, otherwise \end{cases} \quad (13)$$

That is, two objects are correctly related when they share a category if and only if they appear in the same cluster. The Bcubed Precision and Recall and F1 score are defined as follows.

$$Precision\_Bcubed = Avg_e[Avg_{e'.C(e)=C(e')}[correctness(e,e')]] \quad (14)$$

$$Recall\_Bcubed = Avg_e[Avg_{e'.L(e)=L(e')}[correctness(e,e')]] \quad (15)$$

$$F_1 = 2 * \frac{Precision\_Bcubed * Recall\_Bcubed}{Precision\_Bcubed + Recall\_Bcubed} \quad (16)$$

We evaluate the clustering quality in terms of F1 score under varying distance threshold $\rho$, which is shown in Fig.4. From Fig.4, we can observe that F1 score increases gradually as the distance threshold $\rho$ increases from 0.7 to 0.9. F1 score reaches the largest value when $\rho$ is set to be 0.9 and drops slightly after the highest point. Therefore, we can conclude that the clustering quality is the best when $\rho$ is set to be 0.9 and F1 score is 0.68.

We test the scalability of our clustering algorithm in terms of execution time. We set the distance threshold $\rho$ to be 0.9. We evaluate the execution time by varying the volume of data from 50% to 100%. The execution time under different data volume is shown in Fig.5. As we can see, the execution time increase as an increasing in data volume. For example, when 90% data is used for clustering, the execution time is 29.7s.

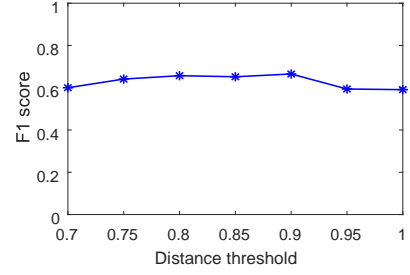

Fig. 4. F1 score for the agglomerative clustering algorithm

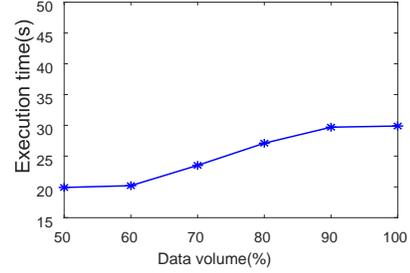

Fig. 5. Scalability of the agglomerative clustering algorithm

*2) Results and Analysis on discovering temporal patterns.* We implement the algorithm 2 TPMiner to extract temporal patterns from Data 2. Three metrics are used to evaluate our TPMiner algorithm: execution time, numbers of patterns, and quality of patterns. We vary the minimum support threshold *minisupport* from 3% to 7% under different volume of data (60% and 80% data volume) to test the execution time of TPMiner. The experimental result is shown in Fig.6. Fig.6 shows the execution time of TPMiner decreases by increasing the minimum support threshold *minisupport*. This is an expected result because the increasing *minisupport* enables TPMiner to generate less temporal patterns in each iteration. Therefore, the search scope of extracting temporal patterns shrinks in the next iteration. In Fig.7, it is clearly that the numbers of extracted temporal patterns decrease significantly as an increase in the minimum support threshold *minisupport*. For example, when *minisupport* is set to be 3%, there are 211 and 342 numbers of temporal patterns found for the 60% and 80% volume of data, respectively.

We may find large numbers of temporal patterns and some of them might be of low quality. It is difficult for users to select these patterns. Therefore, we adapt the metric proposed in [9] to evaluate the quality of temporal patterns in terms of *predictability*. Temporal patterns with high predictability are useful for later activity prediction. Let $Support_{train}(p)$ denote the support of $p$ in the training data. Suppose the length of $p$ is $k$, $Prefix(p)$ is the prefix pattern of $p$ with the length of (k-1). $Support_{train}(Prefix(p))$ is the support of $Prefix(p)$ in the training data. For the pattern $p$, its predictability is denoted as

*Pre* meaning that the pattern *p* could be inferred via its prefix pattern *Prefix*(*p*) with a confidence of *Pre*. The patterns *Prefix*(*p*) and *p* constitute a rule, denoted as < *Prefix*(*p*) → *p* >, which could be used for later activity prediction. The predictability is defined as follows.

$$Pre = \frac{Support_{train}(Prefix(p))}{Support_{train}(p)} \quad (17)$$

We evaluate the numbers of rules generated given different predictability threshold *pre*. We set the minimum support threshold *minisupport* to be 3%. The rationale for setting *minisupport* to be 3% is that temporal patterns with low support might have high predictability and are useful for later activity prediction. We vary the predictability threshold *Pre* from 0.5 to 1. The experimental results are shown in Fig.8. For example, the numbers of rules are 132 when *Pre* is set to be 0.5. It is obvious that the numbers of rules decrease gradually as the predictability threshold *Pre* increases.

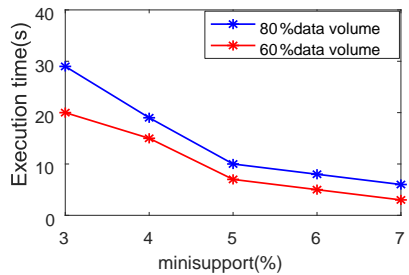

Fig. 6. Scalability of TPMiner algorithm

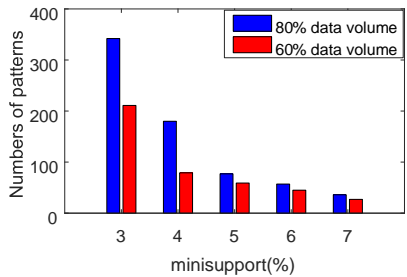

Fig. 7. Numbers of discovered temporal patterns

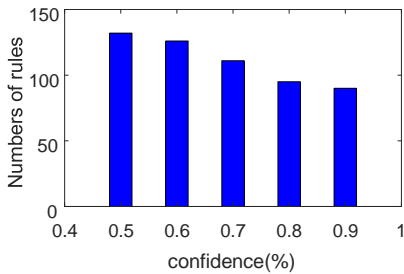

Fig. 8. Numbers of generate rules

## VI. RELATED WORK

### A. Activity discovery

Activity discovery aims to design techniques to discover unknown activity patterns from low level sensor data without any predefined activity models [27]. Sequential pattern mining is an important task and has been extensively studied [6]. The sequential pattern mining is to discover frequent sequences from a set of sequences given a user-specified minimum support threshold [29]. Sequential pattern mining approaches are widely applicable in a variety of domains such as bioinformatics, healthcare, text mining, and consumer's buying patterns [19]. As for the activity discovery, extensions and variants of sequential patterns are proposed to extract activity patterns from time point-based sensor data. In [18], an emerging pattern based approach is proposed to discover complex activity patterns which are used for recognizing sequential, interleaved and concurrent activities. An emerging pattern is a set of items whose frequency changes significantly from dataset to another. For example, a feature vector < (burner), (kitchen) > represents an emerging pattern of "cooking eggs activity" and < (cleanser, plate, water tap), (kitchen) > is an emerging pattern of "washing dishes". The changes between emerging pattern is computed by growth rate. Emerging patterns are those features with large growth rates. A Discontinuous Varied-Order Sequential Miner algorithm is proposed to discover activity patterns from sensor data [51]. This algorithm first discover discontinuous frequent sequences from sensor data. Then, a clustering algorithm is applied to group these discontinuous frequent sequences into clusters. Each cluster corresponds to an activity pattern. An episode pattern discovery approach is proposed to mine regularly occurring device usage patterns [21]. The discovered episodes may be totally or partially ordered and are evaluated based on information theory principles. A frequent periodic pattern mining approach is proposed to find activities that occur frequently and periodically [46]. A variant of the Apriori algorithm is employed to find frequent patterns. Then, a normal distribution method is adopted to find the periods for each frequent pattern.

Sensor activity data may be uncertain due to various factors data incompleteness, noise, and inaccurate measurements [31]. Introducing probability concept is a promising solution to address the data uncertainty problem. In [31], a probabilistic frequency checking algorithm is developed to search the probabilistic frequent spatio-temporal sequential patterns considering time gap. The patterns refers to a set of spatially close objects moving together for certain consecutive timestamps. In [38], a dynamic programming approach is proposed to discover sequential patterns from probabilistic data. Optimization strategies are proposed to reduce candidate generation. In [55], the concept of probabilistic sequential pattern is introduced in discovering surprising periodic patterns. An information gain measure is used to measure the surprise of the periodic patterns. In [56], a probabilistic algorithm is proposed to mine long sequential patterns from noisy data. In [58], the *UPrefixSpan* algorithm is developed to mine probabilistically frequent sequential patterns in large uncertain databases. In our work, we employ a clustering approach to group similar activity data into a cluster and generate the activity pattern associated with each cluster.

*B. Activity recognition*

One hot research topic in smart environment is human activity recognition. The key task of activity recognition is to recognize the ongoing activities from low level data. Typical activity recognition approaches focus on mapping a sequence of low level sensor data to a corresponding activity label. Various activity recognition approaches are designed based on the complexity of the activities. There are various types of activities based the level of complexity. We conceptually categorize activities into two levels: simple activities and complex activities [1]. Simple activities are elementary movement of a person. "walking", "waving hands", "picking a fork" are good examples of simple activities. Complex activities are characterized by the interactions between the resident with one or multiple objects. "sleeping", "bathing", and "watching TV" are typical examples of complex activities. In this paper, we focus on recognizing complex activities.

A number of approaches have been proposed for recognizing activities. According to the different activity data captured by diverse mediums, the activity recognition approaches can be generally classified into two categories. The first category approaches are based on computer vision techniques. They leverage on videos or cameras to capture human activities and exploit computer vision techniques to recognize activity patterns from observations [16]. Much work in human activity recognition has been done in the computer vision community [8]. However, these works mainly focus on recognizing lowlevel simple activities in a controlled environment [43]. The second category approaches are based on sensor networks for collecting human activity data. In these approaches, sensors can be directly attached to persons and objects. Typical sensors can range from switch sensors, infrared motion sensors, pressure sensor, wearable sensors, accelerators, temperature, humidity, and light sensors [17]. In this paper, we are interested in inferring complex activities through monitoring human-object interactions.

There are different approaches facilitate sensor data to recognize the human activities. One category of the approaches considers the activity recognition as a *supervised learning* problem (i.e., classification problem). In the classification, a set of features are extracted from sensor data to build the classifier, and then the classifier is applied to predict the activity label for a sensor event sequence. In [4], a decision tree classifier is trained to recognition simple activities such as siting and walking [54]. In [45], activities are represented as probabilistic event sequences and recognized from the interactions of daily objects. Other sophisticated supervised approaches include support vector machine [24], neural networks [37], and dynamic Bayesian network. When the temporal feature of activities are considered, the temporal classification approaches are developed. In temporal classification, probabilistic-based activity models are employed to recognize the hidden states (i.e., activity labels) from sensor data. The hidden Markov model (HMM) and the conditional random field (CRF) are among the most popular modeling approaches [28]. Supervised methods do not scale well because sensor data need to be annotated manually, which is a very time consuming and laborious task. A different attempt to recognize activities is *unsupervised learning* approach. An emerging pattern mining approach is proposed in [16] to recognize activities. Activities are modeled by emerging patterns which is a feature vector and can describe significant changes between classes of data. Frequent activity patterns are mined from sensor data to build the hidden Markov model for recognizing activities [49]. An episode discovery approach is presented to discover activity patterns within a sequential data stream [22].

*C. Activity prediction*

Activity prediction is to predict the future activities that the resident is likely to perform [40]. According to the definition in machine learning, "prediction" often refers to sequence prediction, where the goal is to predict the next event based on a known limited history of past events [39]. Specifically, the task of sequence prediction is to estimate the next event in a sequence given the events that occur before it [36]. Compared with the activity recognition task which focuses on determining a sequence of activities from a sequence of events, the task of sequence prediction is to forecast which activity (i.e., represented by an event) will occur next.

In [40], the task of activity prediction focuses on predicting both the next activity features and the next activity label. An activity prediction model is proposed using Bayesian networks. A two-step inference process to applied to predict both the next activity features and the next activity label. The start time of the predicted activity is estimated by the continuous normal distribution and outlier detection approaches. In [39], The TEREDA model (TEmporal features and RElations Discovery of Activities) is proposed to discover temporal relations among activities and their start time and duration features. The discovered temporal relations as well as their start time and duration features are used to predict future activities. A frequent pattern mining algorithm is described to find the temporal relations. The expectation maximization (EM) clustering algorithm is employed to construct a normal mixture model for each activity start time and duration. An activity prediction algorithm SPEED ( Sequence Prediction via Enhanced Episode Discovery is proposed to predict resident activity in smart home environment [2]. A set of sequential patterns called episode are first extracted from sensor data. Then, the extracted episodes are transformed into Markov model. A method based on prediction by partial matching (PPM) algorithm is applied to predict the next activity based on the Markov model. In [5], a deep learning approach LSTM (Long Short-Term Memory) is described to predict the next event in a sequence. This approach is demonstrated a better prediction accuracy compared with the SPEED algorithm when the dataset is large. The SPEED algorithm is further improved in [14] by introducing a prefix based mechanism to enhance the tree generation. Another variant of SPEED algorithm called Modified-SPEED algorithm is presented in [35] which includes the time duration and location of the events to decide the next event. In our paper, we focus on predicting next activity label based on temporal patterns among activities.

## VII. CONCLUSION

In this paper, we proposed a novel activity learning framework to develop a prompting system for people's daily life. Our framework consists of three components: activity discovery, activity prediction, and activity recognition. A new clustering algorithm is proposed to extract activity patterns. A new algorithm TPMiner is proposed to discover the temporal relationships among activities. We employ Forward algorithm based on a voting mechanism to recognize the ongoing activity. A rule-based approach is employed to predict the next activity based on the result of activity recognition component. In future, we plan to implement our framework and prototyping the prompting system. We also plan to propose more sophisticated techniques to further improve the performance and scalability of our proposed approach.